\pdfoutput=1

\documentclass[11pt]{article}

\usepackage[final]{acl}

\usepackage{times}
\usepackage{latexsym}
\usepackage{multirow}
\usepackage{amsmath} 
\usepackage{amsmath,amssymb}
\usepackage{multirow}
\usepackage{algorithm}
\usepackage{algorithmic}
\usepackage{booktabs}
\usepackage[T1]{fontenc}

\usepackage[utf8]{inputenc}

\usepackage{microtype}

\usepackage{inconsolata}

\usepackage{graphicx}
\title{Continual-learning for Modelling Low-Resource Languages from Large Language Models}

\author{Santosh Srinath K, Mudit Somani, Varun Reddy Padala, Prajna Devi Upadhyay, Abhijit Das \\ Machine Intelligence Group, Department of CS\&IS,\\ Birla Institute of Technology and Sciences, Pilani, India. \\ Email: abhijit.das@hyderabad.bits-pilani.ac.inl}

\begin{document}
\maketitle

\begin{abstract}
Modelling a language model for a multi-lingual scenario includes several potential challenges, among which catastrophic forgetting is the major challenge. For example, small language models (SLM) built for low-resource languages by adapting large language models (LLMs) pose the challenge of catastrophic forgetting. This work proposes to employ a continual learning strategy using parts-of-speech (POS)-based code-switching along with a replay adapter strategy to mitigate the identified gap of catastrophic forgetting while training SLM from LLM. Experiments conducted on vision language tasks such as visual question answering and language modelling task exhibits the success of the proposed architecture.
\end{abstract}


%

\section{Introduction}


LLMs \citep{chang2024survey} have evolved as a potential aspect of research in the area of artificial intelligence due to their ability to process and interpret both visual and/or textual data for diverse sets of tasks. LLMs often focus on a single modality, but Visual Language Models (VLMs) \citep{Wang_2023_CVPR, MIR-2022-06-193, DBLP:journals/corr/abs-1908-03557,chen2022pali} upscale the context by enabling both visual and textual context at the same time. This capability is particularly important for tasks like image captioning \citep{stefanini2022show, hossain2019comprehensive}, visual question answering (VQA) \citep{lu2023multi, qian2024nuscenes}, and multi-modal content generation \citep{wu2023next, koh2024generating}, where understanding both the image context and text is essential. Additionally, VLMs hold immense potential in areas like assistive technology \citep{de2022artificial,xie2024emerging}, automated content creation \citep{luo2021newsclippings}, and medical diagnostics \citep{delbrouck-etal-2022-vilmedic, boecking2022making}.

Most LLMs and VLMs are trained in high-resource languages like English, limiting their use in a multilingual world. Multilingual LLMs and VLMs are essential to ensure inclusivity, enabling models to understand and generate content in various languages and cultural contexts \citep{ko2022largescale}. Despite the importance of multilingual LLMs and VLMs, a significant challenge remains in modelling these for low-resource languages (LRLs). LRLs are referred to as those languages that have relatively less data available for training the model. There are multiple languages across the globe, and it is estimated that there are around 7000 languages in the world \citep{qin2024multilinguallargelanguagemodel}. Unfortunately, most of the technologies are built in English and only 17\% of the world's population can speak and understand English as per UNESCO \citep{unesco}. 
A major challenge in natural language processing (NLP) is training language models to perform tasks on LRLs. This creates a gap, as speakers of low-resource languages are often excluded from the benefits of VLM technologies. Multilingual language modelling (MLLM) \citep{zhu2023multilingual, geigle2023mblip} is selected as one of the avenues to model LRLs \citep{zhou2021uc2,li-etal-2023-unifying,zhou2021uc2}. 

In MLLM settings, SLMs \citep{schick2021itsjustsizematters} can effectively learn from smaller datasets, via training on LLMs. In other words, LLMs can be fine-tuned to model language models for SLMs without needing the vast amounts of data and resources that large models typically require. Among the existing techniques xMDETR \citep{Wang_2023_CVPR_xMDETR} is a recent work where a code switch (CS) is employed for MLLM for visual SLM. Code switch is a technique that is heavily employed in multilingual language modelling to augment datasets for LRL \citep{Wang_2023_CVPR_xMDETR, m3p}. By definition, code switch is a process of replacing some words in English text in the LRL data following a bilingual dictionary while building the model. M3P \citep{m3p} is a seminal work on code switch used in multilingual multi-modal training. In this work, a pre-trained transformer with altered multilingual, mono-lingual, and multi-modal streams was used. 


There are also simple methods for cross-lingual language modelling, including training an entirely new model from scratch for each new language or training on all languages simultaneously, which can be inefficient and impractical. When faced with a vast array of languages, we use MLLM techniques to maximise learning from data, especially when data  access is limited or short-lived. Continual learning leads to catastrophic forgetting \citep{MCCLOSKEY1989109}, which brings in the challenge of maintaining previously acquired knowledge (stability) while learning new information (plasticity) \citep{NEURIPS2021_ef8446f3}. There are multiple ways of avoiding catastrophic forgetting (CF), such as experience replay \citep{smith2024adaptive}, model expansion \citep{"leitner",mhamdi-etal-2023-cross}, regularisation and distillation-based methods \citep{chen2021overcoming, feng2022overcoming}. To the best of our knowledge, the code switch and catastrophic forgetting have never been addressed in the literature jointly to develop a unified effective model for the multi-lingual scenarios that can get updated with a new set of languages. Moreover, in code switch, syntactical differences are not considered while switching can lead to reduced translation quality, especially for structurally different languages.

On the other hand, one of the model expansion techniques is to integrate adapters to the LLMs. Adapter-based transfer mechanisms \citep{pfeiffer-etal-2020-mad} have proven effective for lightweight domain adaptation and modular training in multilingual settings. While MAD-X delivers strong cross-linguistic performance, existing methods often modify shared adapters or task-specific heads, which risks eroding previously acquired language knowledge.

To address limitations in multilingual retention, we introduce replay adapter a novel architectural component that decouples replay optimization from individual language adapters. This approach maintains strong cross-lingual generalization in continual learning settings with minimal increase ($\sim$ 2 \%) in the overall parameter footprint, thereby enabling scalable and memory-efficient adaptation. Furthermore, to mitigate the dual challenges posed by code switch and multilingual language modeling, we propose a POS-guided code switch strategy integrated with an experience replay mechanism. This combined framework effectively curbs cross-lingual catastrophic forgetting in both vision-language models (VLMs) and large language models (LLMs). The specific contributions of the paper are as follows:

\begin{itemize}
    \item We propose an adapter-based method that enables experience replay for mitigating catastrophic forgetting in a parameter-efficient way.
    
    \item We introduce a novel POS-based code switch strategy designed to preserve syntactic structures during experience replay in continual multilingual learning. 
    
\end{itemize}

\begin{figure*}[h]
    \centering
    \includegraphics[height = 9cm, width=16cm]{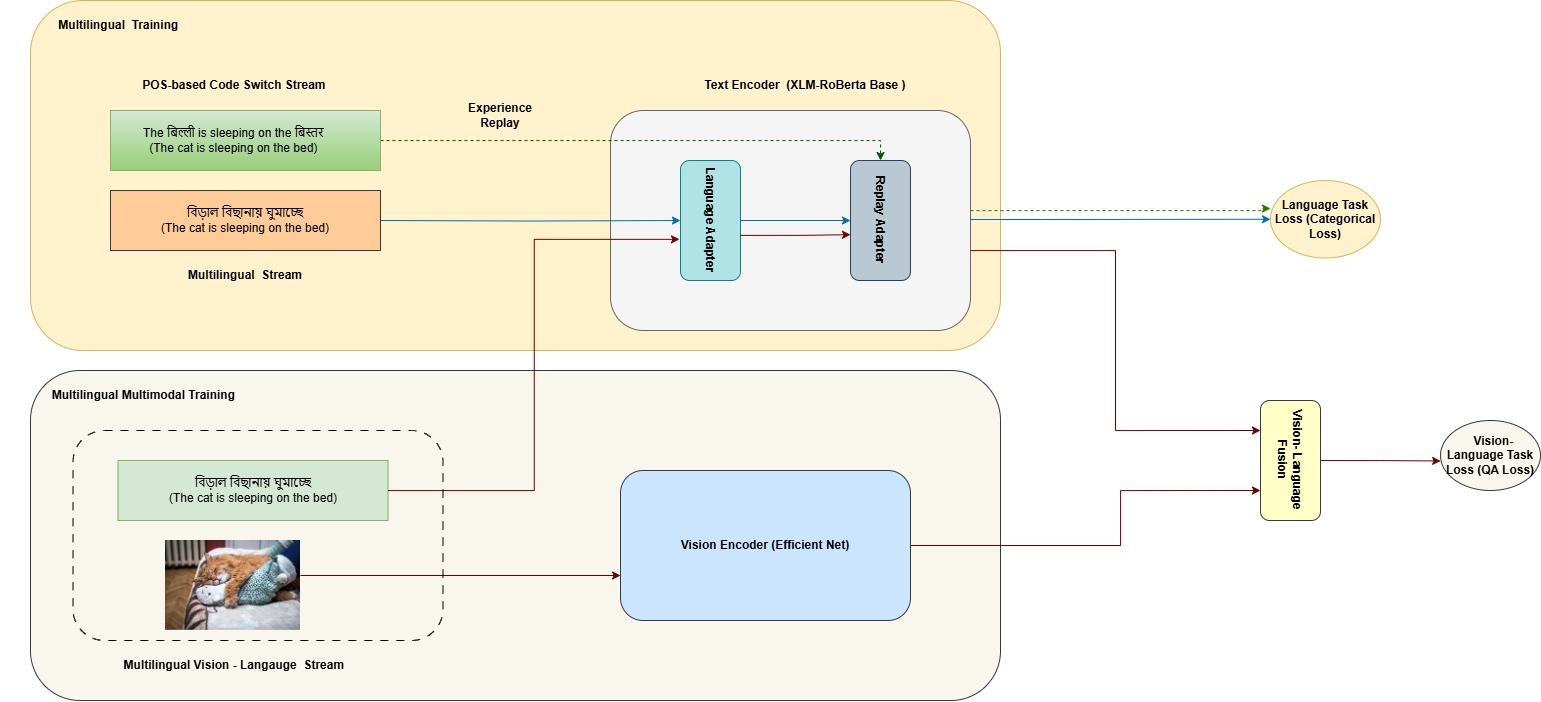}
    \caption{  For multilingual text tasks, the model is trained with multilingual stream (blue)  using both the language adapter corresponding to each language and a shared replay adapter. During experience replay (green), only the replay adapter is updated using a POS-based code switch stream. During multimodal training (red), the image encoder functions as a visual feature extractor, while the language adapter and replay adapter jointly process textual input. The replay adapter is further updated through a POS-based code switch stream }
     
    \label{fig:continual learning fig2}
\end{figure*}

\section{Proposed Methodology}\label{methodology}

\subsection{Overview}
In the proposed multilingual continual learning, the model learns sequentially across languages in multiple phases. In each phase, the model undergoes fine-tuning on a specific language. The model is trained sequentially with different languages in multiple phases. Let $N$ be the total number of phases, equal to the total number of languages for training. Let $l_n$ is the language introduced in phase $p_n$, where
$n \in \{1, 2, \dots, N\}$. In each phase $p_n$, we also replay the previous phase's languages $l_{1:n-1}$ where $n>1$. We evaluate the performance of the model based on a metric $M$ depending on the task. $M_{n,k}$ is the value of a metric for each language $l_k$ computed at each phase $p_n$. However, as training progresses, the model often loses the previously acquired language knowledge, a phenomenon known as catastrophic forgetting. In contrast to previous we proposed to use task-specific replay adapters along with multi-adapter. This adapter-based learning paradigm is especially beneficial for low-resource languages, where training data is scarce \citep{pfeiffer-etal-2020-mad}.

The first phase begins with task-specific training on English. At each phase of continual training, the multilingual text encoder (XLM-RoBERTa Base) is trained using the language adapter of the corresponding language stream and a shared replay adapter.  From the second phase onward, i.e. while training the next language, in addition to training the language, a dedicated replay phase is performed on a fixed frequency basis using a POS-based code switch stream, during which only the replay adapter parameters are updated. In other words, the proposed model adopts a modular text encoder structure, integrating language-specific adapters for each target language and a unified replay adapter to enable experience replay (See Figure 1). Throughout multilingual continual learning, the encoder backbone remains frozen, while both the language-specific adapters and the shared replay adapter are jointly optimised. During replay phases, only the parameters of the replay adapter are updated, preserving the integrity of previously learned representations while mitigating catastrophic forgetting.


\subsection{Proposed POS-Based Code Switch Replay Adapters}
During training, we freeze the base model and only update the adapter weights. For each language $l_i$,  we employ distinct adapter $A_{l_i}$  from adapter set $A$, where  $A = \{A_{l_1},A_{l_2},...A_{l_N}\}$. To enable continual learning across languages, we introduce an additional shared task adapter for replay $A_{replay}$.  However, this design offers a crucial advantage using a separate adapter for replay, enabling more effective retention of prior knowledge, especially compared to relying solely on the classification head, which has limited capacity to preserve language-specific information over time.

Code-switching is a phenomenon in linguistics wherein a multi-lingual speaker alternates between two or more languages in the context of a single conversation. It is also a technique which is used to teach a person a new language by switching words between a language they are already familiar with and the target language; the learner is able to grasp the target language more effectively. In a similar manner, code-switching is also used in training multi-lingual models. 

POS-based code switching mimics this natural linguistic behavior. In addition to grammatical rules, POS-based code switch replacement also preserves the original meaning and context of the sentence \citep{arora-etal-2023-comix}. For instance, a code switch version of the English sentence \textit{The cat
 is sleeping on the bed} with Hindi would be \textit{The billi is sleeping on the bistar}, where the nouns \textit{cat} and \textit{bed} are replaced with \textit{billi} and \textit{bistar} respectively. The code mixed sentence still sounds natural despite the replacement. Therefore, taking into account the syntactic cues can be particularly helpful for training multi-language models on low-resource languages.

 To foster cross-lingual generalization, we integrate a POS-guided code switch strategy during replay in continual training. As mentioned in the Algorithm \ref{algo:cs_strategy}, for each input batch, a configurable fraction of tokens, primarily selected via POS tags, are substituted with equivalents sampled from a target language. Guided by the code switch ratio $\rho$, the subroutine prioritizes token replacements from the specified target part-of-speech (POS) category, where $\rho$ denotes the proportion of tokens to be substituted within each text input  during code-switching. If the available POS-tagged tokens are insufficient to meet the quota ($\alpha_i$), the algorithm supplements them with randomly selected tokens outside the category. This adaptive selection ensures consistent token-level transformation across sentences, irrespective of  POS sparsity.

\vspace{-3mm}
\begin{algorithm}[H]
\caption{POS-Based Code-Switching Subroutine}
\label{algo:cs_strategy}
\begin{algorithmic}[1]
\STATE \textbf{Input:} Batch of sentences $\mathcal{B}$, CS ratio $\rho$, $l_{\text{replay}}$
\STATE \textbf{Output:} Code-switched batch $\mathcal{B}_{cs}$

\STATE Initialize code-switched batch $\mathcal{B}_{cs} \gets \emptyset$

\FOR{each sentence $s_i \in \mathcal{B}$}
    \STATE Compute $\alpha_i = \lceil \rho \cdot |s_i| \rceil$
    \STATE Extract POS-tagged set $\mathcal{P}_c^{(i)}$ of category $c$
    \STATE Define random sets:
    \[
    \mathcal{R}_1^{(i)} \subseteq s_i \setminus \mathcal{P}_c^{(i)}, \quad 
    \mathcal{R}_2^{(i)} \subseteq \mathcal{P}_c^{(i)}
    \]
    \STATE Determine switching targets:
    \[
    \mathcal{S}_{cs}^{(i)} =
    \begin{cases}
        \mathcal{P}_c^{(i)} & \text{if } |\mathcal{P}_c^{(i)}| = \alpha_i \\
        \mathcal{P}_c^{(i)} \cup \mathcal{R}_1^{(i)} & \text{if } |\mathcal{P}_c^{(i)}| < \alpha_i \\
        \mathcal{R}_2^{(i)} & \text{if } |\mathcal{P}_c^{(i)}| > \alpha_i
    \end{cases}
    \]
    \STATE Apply code-switching using $l_{\text{replay}}$ equivalent tokens : $\forall w \in \mathcal{S}_{cs}^{(i)},\ w \mapsto \text{CS}(w)$ \COMMENT{ $\text{CS}(w)$ contains the word after code-switching }
    \STATE Append updated $s_i$ to $\mathcal{B}_{cs}$
\ENDFOR

\STATE \textbf{Return} $\mathcal{B}_{cs}$
\end{algorithmic}
\end{algorithm}
\vspace{-5mm}

 


\subsubsection{Training procedure} \label{section:training}
 The training (Algorithm \ref{algo:training}) begins with English as the anchor language. For each language $l_t$, we initialize a language-specific adapter $A_{l_t}$ while keeping a shared task adapter $A_{\text{replay}}$ persistent across all stages. The base encoder  remains frozen throughout to ensure parameter efficiency and control for catastrophic interference. During training, batches are streamed from the current dataset $\mathcal{D}_{l_t}$, and model updates are applied at the adapter level only. At each batch step, both the language adapter $A_{l_t}$ and the replay adapter $A_{replay}$ are updated.
 However, when the batch index satisfies  $n\bmod f = 0$ and $t>1$, where $n$ denotes the batch counter and $f$ is the replay frequency, a replay event is triggered in which only the replay adapter $A_{replay}$ is updated. Replay involves sampling a batch from the English dataset $\mathcal{D}_{l_1}$, transforming it via a POS-based code switch using $l_p$ equivalent tokens ($ l_p \sim \{l_2, \dots, l_{t}\}$), and updating only the replay adapter with this code-switched input. This targeted strategy ensures retention of prior knowledge and fosters syntactic generalization across diverse typologies without inflating parameter overhead.


\begin{algorithm}[H]
\caption{Continual Training with Returned Code-Switched Replay and Selective Adapter Updates}
\label{algo:training}
\begin{algorithmic}[1]
\STATE \textbf{Input:} Language datasets $\mathcal{D}_{l_1}, \dots, \mathcal{D}_{l_N}$ where $l_1 = \text{English}$, CS ratio $\rho$, replay frequency $f$
\STATE \textbf{Initialize:} Frozen encoder $\theta$, shared task adapter $A_{\text{replay}}$, language adapters $\mathcal{A} = \{A_{l_1}, \dots, A_{l_N}\}$

\FOR{each language $l_t \in \{l_1, \dots, l_N\}$}
    \STATE Load or initialize language adapter $A_{l_t}$
    \STATE Set batch counter $n \gets 0$

    \FOR{each batch $\mathcal{B}_t \in \mathcal{D}_{l_t}$}
        \STATE Increment counter: $n \gets n + 1$

        \IF{$t > 1$ and $n \bmod f = 0$}
            \STATE Sample replay batch $\mathcal{B}_r \sim  \mathcal{D}_{l_1}$
            \STATE Sample target language for code-switching $l_{\text{replay}} \sim \{l_2, \ldots, l_t\}$
            \STATE Call Code-Switching Subroutine: $\mathcal{B}_{cs} \gets \text{CS-Subroutine}(\mathcal{B}_r, \rho, l_{\text{replay}})$
            \STATE \textbf{Update only} replay adapter $A_{\text{replay}}$ using $\mathcal{B}_{cs}$
        \ELSE
            \STATE Update language adapter $A_{l_t}$ using $\mathcal{B}_t$
            \STATE Update replay adapter $A_{\text{replay}}$ using $\mathcal{B}_t$
        \ENDIF
    \ENDFOR
\ENDFOR
\end{algorithmic}
\end{algorithm}

\section{Experimental Setup}
In this section, we will explain the dataset employed for experiments, experimental setting, results and analysis. 

\subsection{Dataset}
We evaluate proposed method with datasets on diverse tasks of different languages and modalities to understand its effectiveness in mitigation of catastrophic forgetting.  
For MLVLM experiments, we leverage xGQA \citep{pfeiffer2021xgqa} to assess cross-lingual multimodal alignment under syntactic perturbations. Its structured multilingual vision question answering format, spanning seven typologically diverse languages, facilitates controlled evaluation of POS-based replay and code-switching strategies, thereby reflecting knowledge retention dynamics and modality-specific interference in continual learning

 For MLLM experiments, we use the MTOP \citep{li2020mtop} and PAXQA \citep{li-callison-burch-2023-paxqa} datasets. MTOP is designed for multilingual intent classification that covers 117 intents. Its parallel annotations is useful for probing syntactic retention, semantic drift in multilingual continual learning setups. PAXQA, a multilingual QA dataset spanning languages such as English, Arabic, Russian, and Chinese, is leveraged to draw comparative insights against multimodal question answering benchmarks. The dataset is generated using  parallel corpora like GALE \citep{ldc2008t09}, NC \citep{barrault2020wmt}, and UN \citep{ziemski2016un}. Its diverse sources and lexically-constrained translation is useful for continual learning studies. For the experiments, languages are selected from a common source corpus to maintain consistency across training data.

For evaluation on the MTOP and xGQA datasets, missing language variants were supplemented via automatic translation using the NLLB \citep{nllb2022}. 

\subsection{Code-switch Lexicon Preparation}
We prepare bilingual lexicon using MUSE \citep{conneau-2020-unsupervised,lample2017unsupervised} bilingual embeddings, which are aligned with the vocabulary setup in XMDETR \citep{Wang_2023_CVPR_xMDETR}. To ensure token-level consistency, multiword expressions are excluded, and named entities are transliterated to preserve phonetic fidelity across scripts.

\subsection{Metrics}

\label{metrics}
The Table \ref{table:dataset} summarizes the primary evaluation metric $M$ used for each dataset to monitor model performance at every epoch. To assess catastrophic forgetting across all datasets, we rely on the average accuracy \citep{chaudhry2018riemannian}.
%
%
\[
AA = \sum_{k = 1}^{N}M_{N,k}
\]
where $M_{n,k}$ is the value of a metric for language $\ell_k$ evaluated at phase $p_n$ and $N$ is the total number of languages in the continual training sequence.
\begin{table}[t]
\small
\begin{tabular}{cccc}
\hline
Dataset &Modality &Task & Metric ($M$) \\
\hline
MTOP & Text&Intent cls &Accuracy \\
PAXQA &Text &QA  &Exact Match  \\
XGQA &Text+Image &VQA &Accuracy  \\
\hline
\end{tabular}
\caption {This table outlines the datasets used for evaluation. MTOP involves intent classification (Intent cls), GQA focuses on vision question answering (VQA), and PAXQA covers the question answering (QA) task. The exact match metric \citep{lewis-etal-2020-mlqa} is a strict metric, useful for measuring precision in extractive QA tasks.
 } 
\label{table:dataset}
\end{table}


\begin{table*}[t] 
\small
\centering
\setlength{\tabcolsep}{2.3pt}

\begin{tabular}{cccccccccc}
\hline

\hline
& \multicolumn{2}{c}{\underline{Lower Bound}}& \multicolumn{6}{c}{\underline{POS-based CS}} & \underline{Upper Bound}   \\ 
 Language Sequence & {No Replay}  & Random CS  &   {NOUN } &  {VERB} &  {ADJ} &{ADV} &  {PROPN} &  {INTJ} &   \\ 
 \toprule
 MTOP dataset \\
\toprule               
 $EN \rightarrow FR \rightarrow ES$     & 84.57  & 89.55  & 89.62 & 89.52 & \textbf{89.69} & 89.4 & 89.55 & 89.65  & 92.38   \\

$EN \rightarrow ES \rightarrow FR$ & 75.33   & 89.87  &90.05   & 90.06 & 90.03  & \textbf{90.07}  &90.03  &90.01  & 92.38      \\

$EN \rightarrow TH \rightarrow KO$   & 84.19  &  89.86  & 89.98 &90.01  & \textbf{90.08} & 89.92 &90.03  &  90.04 & 91.63   \\

$EN \rightarrow DE \rightarrow FR $   & 86.7 & 90.52    & 90.4 & 90.59 & 90.6 & 90.59 & 90.60 & \textbf{90.63} & 92.34 \\

$EN \rightarrow HI \rightarrow BN$     & 91.67   & 91.49  & 91.54 & 91.59 & \textbf{91.80} &  91.57 & 91.60 & 91.61 & 92.18 \\
\midrule
Average & 84.49 & 90.25 & 90.31 & 90.35 & \textbf{90.44} & 90.31 & 90.36 & 90.38 & 92.18 \\ 
\toprule
XGQA dataset \\
\toprule

  $EN \rightarrow PT \rightarrow ES $ & 29.11 & 33.31 & 31.27 & 34.82 & 33.23 & 32.96 & \textbf{35.05} & 33.73 &  36.83 \\
 $EN \rightarrow HI \rightarrow BN$ & 27.76 & 27.63 &  26.20 &  27.65 & 26.89 & \textbf{28.26} & 27.19 & 27.70 &  37.61\\
 $EN \rightarrow KO \rightarrow ID$ & 28.23 & 29.35 & 28.13 & \textbf{30.00} & 29.56 & 27.84 & 29.14 & 29.21 &  35.42\\
\midrule
Average & 28.36 & 30.09 & 28.53 & \textbf{30.82} & 29.89 & 29.68 & 30.45 & 30.21 & 36.62 \\
\toprule
PAXQA dataset \\
\toprule
  $EN \rightarrow AR \rightarrow ZH  $ & 61.68 & 64.33 & \textbf{67.49} & 64.33 & 65.13 & 62.19 & 64.58 & 64.46 &  70.53 \\

$EN \rightarrow RU \rightarrow ZH $ & 78.20 &77.25&77.66&78.12& 76.35 & 77.17& \textbf{78.23} & 76.38 & 79.86 \\ 
$EN \rightarrow ZH \rightarrow RU  $  &77.45&78.41&78.79 & \textbf{79.36} &77.98 &76.64 &77.28&76.99 & 79.86 \\ 
$EN \rightarrow RU \rightarrow AR $ & 74.48 & 72.39 & 72.65 & 71.87 & 72.65 & \textbf{75.26} & 72.91 & 72.91 & 78.12 \\
\midrule
Average & 73.09 & 73.09 & \textbf{74.14} & 73.42 & 73.02 & 72.81 & 73.25 & 72.68 & 77.09 \\
\bottomrule

\end{tabular}
\\[5pt]
\caption {This table presents the results of continual training on MTOP, PAXQA and XGQA datasets for various language sequences using average accuracy (Section \ref{metrics}). Evaluation of POS-based code switch is focused on POS categories like noun (NOUN), verb (VERB), adjective (ADJ), adverb (ADV), proper noun (PROPN), interjection (INTJ). Details of lower bound and upper bound are mentioned in the Section \ref{base_lines}. The \textbf{bold} values indicate the highest average accuracy for the corresponding language sequence. A higher average accuracy signifies better cross-lingual knowledge retention.}
\label{table:main_results}
\end{table*}

\subsection{Implementation details}
  For all the experiments, we use XLM-RoBERTa Base \citep{conneau-2020-unsupervised} as the text encoder with 12 layers as a trained multilingual transformer-based encoder model. Instead of fine-tuning the full multilingual text encoder, we employ MAD-X adapter modules \citep{pfeiffer-etal-2020-mad} within each Transformer layer and optimize only the adapter parameters, thereby preserving the pretrained encoder's linguistic generalization capabilities. If specific pretrained language adapters are not available, we train the adapter from scratch using the task specific training data. For the replay adapter, we use the same type of adapter as the language adapter \citep{DBLP:journals/corr/abs-2007-07779}.  For finetuning the adapters, we employ the Adam optimizer setting the learning rate to 3e-5 with a batch size of 16 for all experiments on different datasets. Across all experiments, we fix the code-switching ratio ($\rho$) at 0.5 to maintain a balanced distribution between original and switched tokens. Experience replay is integrated into continual training at regular intervals, specifically every 10th batch ($f$ = 10). In each phase of language learning, for the MTOP and PAXQA dataset, we ran a maximum of 10 epochs. Based on our experiments, we have not seen significant improvement in performance beyond 10 epochs for each task.

  To train the XGQA dataset, we finetune using both the language adapters and replay adapter integrated to  multilingual text encoder (XLM-RoBERTa Base) of the XMDETR architecture based on proposed method, while keeping the remaining components of the XMDETR architecture frozen. We train the model for 60K steps with a batch size of 16 and a learning rate of 1e-4.
\subsection{Baselines}
\label{base_lines}
The lower-bound experiments are divided into two types: (1) \textit{no replay} - where prior language data is excluded from experience replay, and (2) \textit{random CS} - where random words are substituted with previous languages during replay. For upper-bound experiments, we employ joint training across all languages rather than sequentially (continual learning).  
\subsection{Results}
We conduct experiments on the MTOP, PAXQA, and XGQA datasets across thirteen languages: English (EN), Chinese (ZH), German (DE), French (FR), Russian (RU), Arabic (AR), Spanish (ES), Portuguese (PT), Korean (KO), Indonesian (ID), Thai (TH), Hindi (HI), and Bengali (BN). Detailed results are presented in Table~\ref{table:main_results}. In our experimental setup, we target open-class categories from the Universal POS tagset \citep{jurafsky2025speech} due to their consistent replay effectiveness and minimal cross-lingual interference, as supported by prior findings \citep{de-vries-etal-2022-make}.

\noindent\textbf{Results on MTOP}
Across diverse language sequences, POS-based code switch driven by adjectives consistently outperform other parts of speech, even though nouns and verbs tend to encode more semantic information compared to other POS categories. Although interjections generally encode limited information, their inclusion in code-switching proves to be notably impactful in enhancing the effectiveness of experience replay. We can also observe that POS-based code switch outperforms random code switch for all language sequences. Another important observation is that the language sequences that are not syntactically similar ($EN \rightarrow HI \rightarrow BN$ and $EN \rightarrow TH \rightarrow KO$) to English (base language for code switch) still benefit POS-based code switch. The Hindi and Bengali are similar languages with high lexical sharing that helps to retain knowledge even without any form of replay. 

\noindent\textbf{Results on XGQA}
\label{sec:analysis_vlm}
POS-based code switch consistently outperforms random code switch across all evaluated language sequences. Notably, the $EN \rightarrow PT \rightarrow ES$ sequence yields the most significant gains from POS-based code switch, likely due to the high linguistic proximity between Portuguese and Spanish, with performance approaching the upper bound
. In contrast, sequences such as $EN \rightarrow HI \rightarrow BN$ and $EN \rightarrow KO \rightarrow ID$  exhibit more modest gains, likely due to the challenges posed by low-resource language status and the increased complexity of multimodal tasks inherent to XGQA.

\noindent\textbf{Results on PAXQA}
The PAXQA dataset comprises high-resource languages. Across all evaluated sequences, POS-based code switch consistently outperforms random substitution. We also observe that order of languages in language sequences plays an important role in knowledge retention. Although the languages are the same in the sequences $EN \rightarrow RU \rightarrow ZH$ and $EN \rightarrow ZH \rightarrow RU$, the latter sequence has better knowledge retention due to smoother syntactic transition of the languages in continual learning and also probably due to code switch compatibility with English. Though the sequence $EN \rightarrow AR \rightarrow ZH$ yields a lower average accuracy relative to other paths possibly due to Arabic-induced syntactic interference, POS-based code switch still outperform random code switch.

\begin{table}[t]
\scriptsize
\begin{tabular}{cccc}
\toprule   
Language Sequence & POS-based CS& Random& Sample Size \\

\toprule
\multirow{5}{*}{$EN \rightarrow HI \rightarrow BN$}&\textbf{91.55}& 91.44&	10\% \\
&\textbf{91.70} &91.59&	30\% \\
&\textbf{91.53}& 91.48&	50\% \\
&91.61& \textbf{91.64}&	70\% \\
&\textbf{91.8}& 91.49&	100\% \\
\midrule
\multirow{5}{*}{$EN \rightarrow TH \rightarrow KO$}&\textbf{89.68} & 89.65 & 10\% \\
&\textbf{89.80} & 89.75 & 30\% \\
&89.90 & \textbf{90.20} & 50\% \\
&\textbf{89.74} & 89.66 & 70\% \\
&\textbf{90.08} & 89.86 & 100\% \\
\bottomrule
\end{tabular}
\caption{Average accuracy scores on the MTOP dataset using different memory replay sample sizes. \textbf{Bold} indicates the top score per language sequence.}
\label{table:memory_size}
\end{table}

\subsection{Comparison with  Learning without Forgetting (LWF) and  Elastic Weight Consolidation
 (EWC)}
 LWF \citep{li2016learning}  mitigates catastrophic forgetting by distilling knowledge from previous tasks without requiring access to old data. It preserves prior task performance by aligning logits between the current and frozen models during continual updates. EWC \citep{kirkpatrick2017overcoming} is a regularization method that penalizes changes to parameters that are important for previously learned tasks by using the Fisher Information Matrix to estimate their importance. To make it comparable with the proposed approach, we use adapters for training each language task. As shown in Table \ref{table:continual}, the proposed method exhibits consistent robustness across languages, outperforming prior approaches particularly in low-resource scenarios.

\subsection{Ablation Studies}
\label{section:ablation}
\textbf{Impact of Memory Size in Experience Replay}
The Table \ref{table:main_results} represents the results of continual learning by replaying code switch with English as the base language. In this section, we explore the impact of the memory sample size in experience replay on both random and POS-based code switch strategies. In the Table \ref{table:memory_size}, we observe that as the memory sample size increases in the experience replay, POS-based code switch performs better than the random code switch.  
\subsection{Effectiveness of Replay Adapter}
The replay adapter facilitates the preservation of cross-lingual knowledge during experience replay. Without it, the burden of knowledge retention shifts entirely to the classification head. The Table \ref{table:replay_adapter_vertical} reports the effectiveness of the replay adapter in mitigation of catastrophic forgetting.  
\begin{table}[t]
\scriptsize
\centering
\begin{tabular}{lcc}
\toprule
& Random CS & POS-based CS \\
\midrule
\multicolumn{3}{l}{\textbf{Sequence: $EN \rightarrow FR \rightarrow ES$}} \\
\quad Without Replay Adapter & 62.31 & 62.44 \\
\quad With Replay Adapter    & 89.55 & \textbf{89.69} \\
\midrule
\multicolumn{3}{l}{\textbf{Sequence: $EN \rightarrow ES \rightarrow FR$}} \\
\quad Without Replay Adapter & 81.56 & 81.57 \\
\quad With Replay Adapter    & 89.87 & \textbf{90.07} \\
\midrule
\multicolumn{3}{l}{\textbf{Sequence: $EN \rightarrow TH \rightarrow KO$}} \\
\quad Without Replay Adapter & 88.60 & 89.23 \\
\quad With Replay Adapter    & 89.86 & \textbf{90.80} \\
\midrule
\multicolumn{3}{l}{\textbf{Sequence: $EN \rightarrow DE \rightarrow FR$}} \\
\quad Without Replay Adapter & 88.40 & 88.10 \\
\quad With Replay Adapter    & 90.52 & \textbf{90.63} \\
\midrule
\multicolumn{3}{l}{\textbf{Sequence: $EN \rightarrow HI \rightarrow BN$}} \\
\quad Without Replay Adapter & 91.33 & 91.33 \\
\quad With Replay Adapter    & 91.49 & \textbf{91.80} \\
\bottomrule
\end{tabular}
\caption{Average accuracy scores on the MTOP dataset, comparing random and POS-based code switch with and without a replay adapter. \textbf{Bold} indicates the highest score per sequence.}
\label{table:replay_adapter_vertical}
\end{table}

\begin{table}[t]
\scriptsize
\begin{tabular}{ccccc}
\toprule

Language Sequence & LWF & EWC & Proposed Method \\
\toprule
MTOP dataset \\
\toprule
$EN \rightarrow FR \rightarrow ES$ & 88.94& \textbf{90.88} & 89.69 \\ 
$EN \rightarrow ES \rightarrow FR$ &88.73& \textbf{91.46} & 90.07\\
$EN \rightarrow TH \rightarrow KO$ &87.89& 89.85 & \textbf{90.08} \\
$EN \rightarrow DE \rightarrow FR$  &89.33& \textbf{91.49} & 90.63\\
$EN \rightarrow HI \rightarrow BN$  &78.67 & 90.32 & \textbf{91.8} \\
\toprule
PAXQA dataset \\
\toprule
$EN \rightarrow AR \rightarrow ZH$  & 51.71& 54.11 & \textbf{67.49} \\
$EN \rightarrow RU \rightarrow ZH$ & 60.88& 63.44& \textbf{78.23} \\
$EN \rightarrow ZH \rightarrow RU$ &60.5&60.55& \textbf{79.36}\\
$EN \rightarrow RU \rightarrow AR$ & 61.26 &62.37 & \textbf{75.26}\\
\toprule
\end{tabular}

\caption {Comparison of proposed method with LWF and EWC. \textbf{Bold} indicates the highest average accuracy per sequence.}
\label{table:continual}
\end{table}

\subsection{Detailed analysis}
\textbf{POS Tag Frequency and average accuracy Correlation}
 In this section, we discuss the relation between POS frequency distribution and the average accuracy mentioned in  Table \ref{table:main_results}. The correlation analysis follows a two-step approach. We aggregate the POS frequency distributions across languages by computing their average, followed by calculating the Pearson correlation with the corresponding average accuracy of POS category.  For example, we calculate the average frequency of verbs of all languages in the sequence across all language sequences and then correlate with average accuracy values of all language sequences under verb category as mentioned in the Table \ref{table:main_results}. As shown in the Table \ref{table:correlation}, there exists a strong negative correlation between the aggregated POS frequency distribution across all languages $(l_1, l_2,l_3)$ in the sequence and its corresponding average accuracy in the MTOP dataset. The frequency distribution of adjectives demonstrates the highest correlation with the average accuracy, a pattern that remains consistent with the overall average accuracy across all POS categories. In PAXQA dataset, the optimal POS categories differ across language sequences, likely because POS frequency shows little correlation with the choice of optimal category. 
         
\begin{table}[t]
\tiny
\begin{tabular}{ccccccc}
\toprule
Langs & NOUN & VERB & ADJ & ADV & PROPN & INTJ \\
\toprule
MTOP dataset \\
\toprule

$l_2$ & -0.76 &	-0.57&	\textbf{-0.77} &	-0.35 &	-0.57 &	-0.26 \\
$l_3$ & -0.54 &	-0.37 &	-\textbf{0.69} &	-0.35 &	-0.61 &	-0.54
\\

$(l_2,l_3)$ & -0.64 &	-0.48 &	\textbf{-0.89} &	-0.41 &	-0.59 &	-0.64 \\
$(l_1,l_2)$ & -0.76 &	-0.58 &	\textbf{-0.76} &	-0.35& 	-0.57&	-0.26 \\

$(l_1,l_2,l_3)$ &-0.64 &	-0.48 &	\textbf{-0.89} &	-0.41 &	-0.60 & -0.67 \\

\toprule
PAXQA dataset \\
\toprule

$l_2$ &	0.02 &	0.78 &	-0.16&	\textbf{0.93}&	0.89&	-0.86\\
$l_3$	&0.37&	-0.07&	\textbf{0.57}&	0.14&	0.36&	-0.36 \\
$(l_1,l_2)$&	0.31&	0.83&	0.35&	\textbf{0.95}&	0.93&	-0.92 \\
$(l_2,l_3)$	&0.77&	0.65&	0.78&	0.80&	0.93&	\textbf{-0.99} \\
$(l_1,l_2,l_3)$ &0.75&	0.71&	\textbf{0.99}&	0.79&	0.87&	\textbf{-0.99} \\
\toprule
\end{tabular}
\caption{This table shows the Pearson correlation between POS frequency in combined training data and average accuracy across language sequences and POS categories. The \textit{Langs} column lists languages used for aggregated POS distribution, and \textbf{bold} values mark the strongest negative correlations.
}
\label{table:correlation}
\end{table}
\textbf{Attention Weights Analysis}
To further analyze the behavior of POS-based code-switch in comparison with random code switch, we quantified the model's internal processing using the following metrics:
\begin{itemize}

\item Attention Entropy : This metric measures the distribution of the model's focus \cite{clark2019bert}. Higher entropy values indicate that the model is relying on the broader sentence context to resolve meaning, rather than fixating on specific lexical tokens (keywords).

\item Attention Mass : Interpreted through the lens of sub-words \cite{rust2021tokenizer}, this metric analyzes the computational load. A higher attention mass signifies increased token fragmentation, demonstrating that the code-switched anchors impose a greater processing requirement on the encoder.
\end{itemize}

We conducted experiments and performed statistical analysis on the MTOP dataset using a model trained on the English → Hindi → Bengali sequence, and measured attention metrics for Hindi. First, the higher Attention Entropy observed in POS-based code-switching suggests that the model moves beyond simple keyword shortcuts (McCoy et al., 2019) and instead distributes its focus across the full sentence context to verify meaning \cite{clark2019bert}. Second, regarding Attention Mass, our results show that POS-based code-switching produces higher values, indicating that it introduces fragmented, high-density tokens which require greater processing effort from the model (Rust et al., 2021). Overall, our results show that POS‑guided code switch provides a stronger training signal than random code‑switching, pushing the model to focus on sentence structure.

\begin{table}[h]
\scriptsize
\begin{tabular}{lcc}
\hline
\textbf{Metric} & \textbf{Attention Entropy} & \textbf{Attention Mass} \\
\hline
Random code switch & $2.09$ & $12.43 $ \\
POS-based code switch & $2.18 $ & $13.31 $ \\
\hline
\end{tabular}
\caption{Comparison of Attention Metrics by Code Switch Type}
\label{tab:attention_metrics}
\end{table}



\begin{figure}[ht]
    
    \centering
    \includegraphics[width=0.48\textwidth]{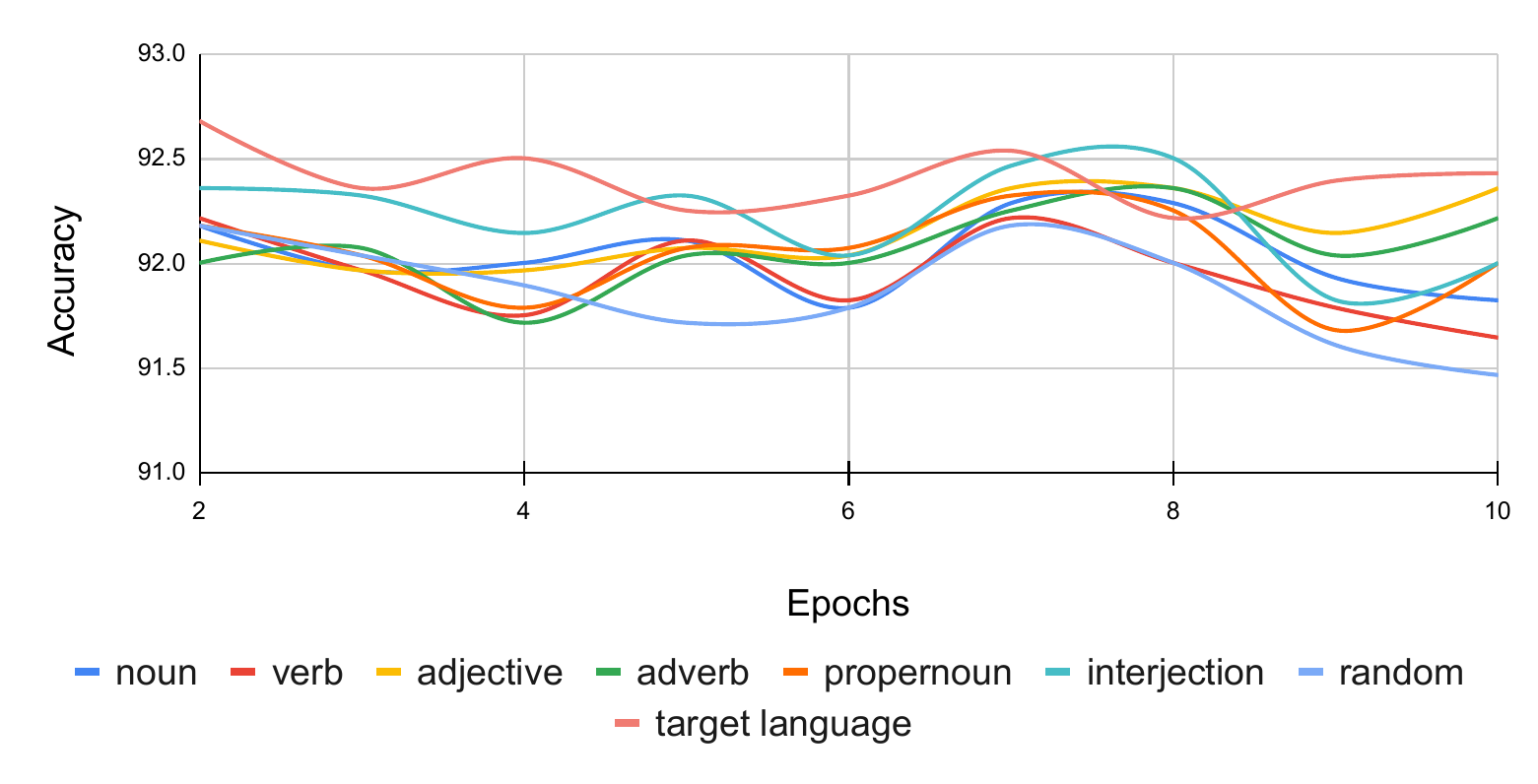}
    \caption{ Cross-lingual retention curve for Hindi during Bengali training in the $EN \rightarrow HI \rightarrow BN$ sequence, evaluated after each epoch to assess transfer stability}
    \label{fig:en_hi_gn}
\end{figure}

\noindent\textbf{Language Specific Forgetting Analysis}
During training on Bengali in the sequence $EN \rightarrow HI \rightarrow BN$, we incorporate replay from Hindi and track Hindi accuracy after each epoch under various replay strategies, as shown in
Figure \ref{fig:en_hi_gn}. Notably, POS-based code switch helps retain cross-lingual knowledge more effectively than random code switch. Additionally, there is not much forgetting of Hindi knowledge. This may be attributed to the linguistic proximity between Hindi and Bengali that share substantial lexical overlap.

\begin{figure}[ht]
    \centering
    \includegraphics[width=0.48\textwidth]{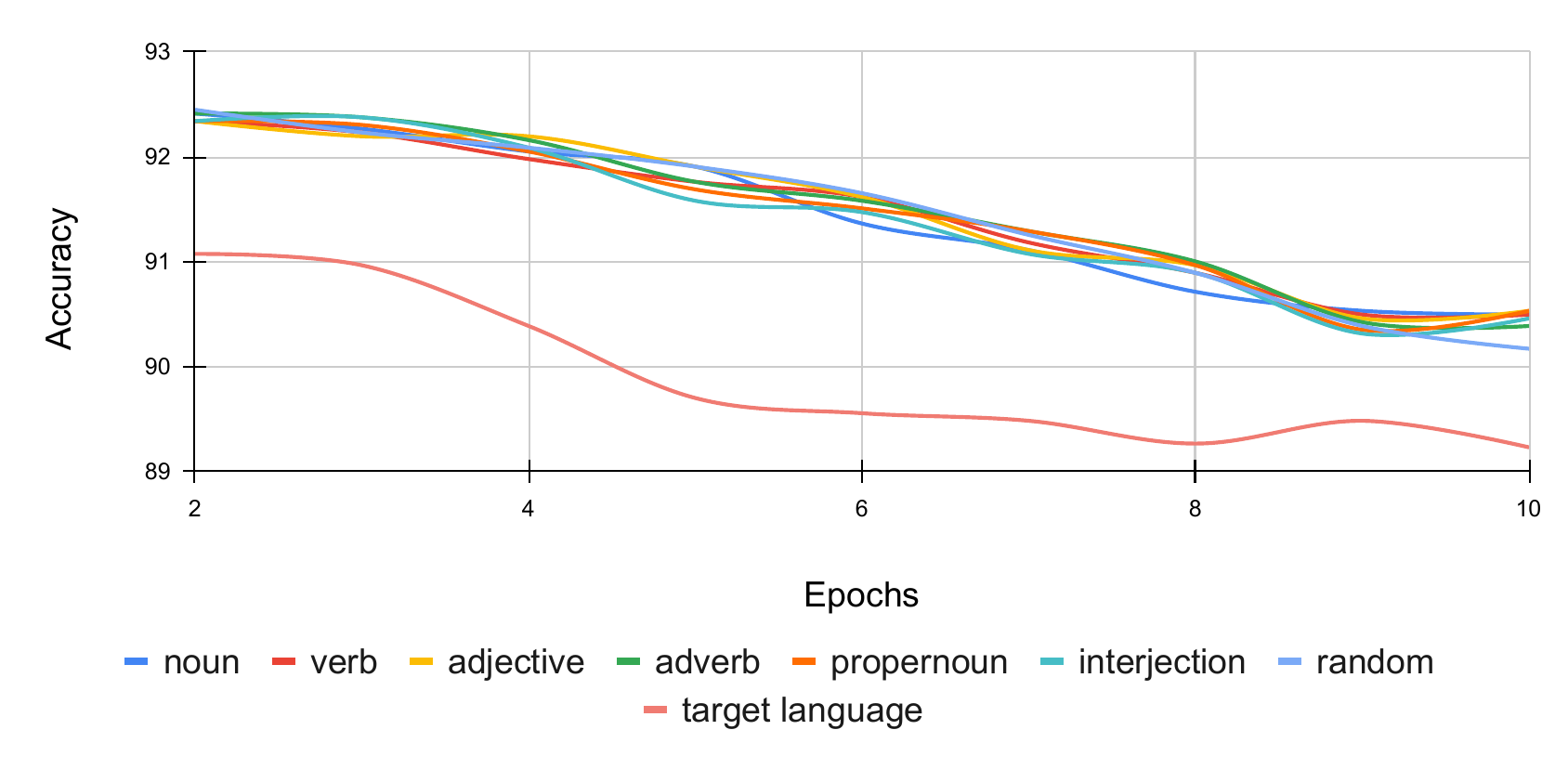}
    \caption{Cross-lingual retention curve for Thai during Korean training in the $EN \rightarrow TH \rightarrow KO$ sequence, evaluated after each epoch to assess transfer stability}
    \label{fig:en_th_ko}
\end{figure}

In contrast, Figure \ref{fig:en_th_ko} shows that although Thai knowledge degrades over time during training on Korean, code switch based replay still outperforms replay using the target language alone. This can be attributed to the linguistic divergence of the Korean and Thai languages.


\begin{table}[t]
\scriptsize
\begin{tabular}{ccccccc}
\toprule
Layer & Seq 1 & Seq 2 & Seq 3 & Seq 4 & Seq 5 & Avg \\
\toprule
Layer 1 &	0 &	0 & 0 & 0 & 0 & 0 \\

Layer 2 &	-1.79 &	-0.5 & -0.87 & -1.27 & \underline{-5.26} & \underline{-1.93} \\

Layer 3 &	-0.14 &	+0.29 & +4.13 & -1.37 & -2.76 & +0.03 \\

Layer 4 &	+0.57 & -0.53 & -0.04 & -0.33 & +0.2 & -0.02 \\

Layer 5 &	-1.36 & +0.63 & \underline{-2.50} & -1.24 & -0.62 & -1.01 \\

Layer 6 &	\underline{-3.45} & \underline{-1.13} & +0.04 & +3.24 & \textbf{+0.59} & -0.14 \\

Layer 7 &	-0.82 & -0.29 & +0.03 & +1.43 & -2.19 & -0.36 \\

Layer 8 &	-1.51 & -0.47 & -0.25 & -2.87 & -1.89 & -1.39 \\

Layer 9 &	+0.61 & -1.07 & +1.49 & -1.60 & -0.09 & -0.13 \\

Layer 10 &	+0.50 & \textbf{+3.60} & \textbf{+9.84} & \underline{-8.14} & -2.62 & +0.63 \\

Layer 11 &	\textbf{+1.65} & +6.80 & +0.91 & \textbf{+20.35} & -4.59 & \textbf{+5.02} \\

Layer 12 &	+1.11 & -0.44 & +3.98 & +14.28 & -0.51 & +3.68 \\
\bottomrule
\end{tabular}

    \caption{ Layer-wise forgetting is quantified in this table  via accuracy shifts in the second language $(l_2)$ across phases, capturing retention (positive) and erosion (negative). \textbf{Bold} values denote peak retention per sequence, while \underline{underlined} values indicate pronounced forgetting.} 
     \label{table:layer_probing}
\end{table}

\noindent \textbf{Forgetting Analysis using Layer Probing}

To evaluate linguistic retention throughout continual learning, we extract language-specific embeddings from replay adapters at each encoder layer. These layer-wise representations serve as input to lightweight classifiers trained for task-specific predictions \citep{alain2016understanding}. Comparing classifier accuracies across layers and phases allows us to track the trajectory of knowledge retention over time. We performed experiments on MTOP dataset for the language sequences $EN \rightarrow HI \rightarrow  BN$ (Seq1), $EN \rightarrow FR \rightarrow  ES$ (Seq2), $EN \rightarrow TH \rightarrow KO$ (Seq3),  $EN \rightarrow ES \rightarrow  FR$ (Seq4), $EN \rightarrow DE \rightarrow  FR$ (Seq5). The Table \ref{table:layer_probing} reveals that forgetting predominantly occurs at lower encoder layers, while the upper layers exhibit signs of positive backward transfer \citep{diaz2018dont}, indicating reinforcement of earlier linguistic knowledge through replay adapter.

\textbf{Experiments on Decoder-only model}
To evaluate the proposed method, we implemented it using LLAMA2 \cite{touvron2023llama}. For architectural parity with XLM-Roberta, we applied deep pruning to retain only 12 layers. We prune LLAMA2 using a representative subset of layers: early (layer 0-3) middle (layers 8, 14, 20, 26) and final (layers 28–31). This selection captures hierarchical representations while minimizing computational overhead. This configuration balances representational diversity while reducing computational overhead.
 We reformulated the MTOP intent classification task as text generation using the template shown in Table \ref{tab:prompt_structure}. We conducted hyperparameter tuning over POS categories and selected the optimal category for code-switching.  Table \ref{table:decoder_results} demonstrates two distinct effects: POS-based switching improves generation quality over random switching, and the replay adapter enhances continual learning by reducing forgetting. 

\textbf{Comparison with Other Machine Translation Models}
We translated from English version of data to Bengali for MTOP dataset, leveraging English as a high‑resource language \cite{Kann2022HighResourceMT}, using IndicTrans2‑en‑indic‑1B model \cite{gala2023indictrans} and compared results against NLLB for the sequence  $EN \rightarrow HI \rightarrow  BN$. We observe similar results across both NLLB and IndicTrans2 in Table \ref{tab:vertical-replay-pos}, confirming the robustness of our findings.
\begin{table}[ht]
\centering
\footnotesize
\begin{tabular}{l|c|c}
\hline
{Metric}           & {NLLB} & {IndicTrans2} \\
\hline
No Replay                 & 91.67         & 90.00 \\
Random                    & 91.49         & 90.70 \\
NOUN                      & 91.54         & 91.46 \\
VERB                      & 91.59         & 91.45 \\
ADJ                       & 91.80         & 91.54 \\
ADV                       & 91.57         & 91.07 \\
PROPN                     & 91.60         & 91.05 \\
INTJ                      & 91.61         & 91.34 \\
Joint Training            & 92.18         & 92.03 \\
\hline
\end{tabular}
\caption{Vertical comparison of NLLB and IndicTrans2 across replay strategies and POS categories.}
\label{tab:vertical-replay-pos}
\end{table}

\begin{table}[t]
\scriptsize
\begin{tabular}{cccc}
\toprule
Language Sequence & No Replay & Random CS & POS-based CS  \\
\toprule

$EN \rightarrow FR \rightarrow ES$  & \textbf{74.11} & 73.98& 73.8\\ 
$EN \rightarrow ES \rightarrow FR$ & 74.57 & \textbf{74.72} & 74.57 \\
$EN \rightarrow TH \rightarrow KO$ & 58.16 & \textbf{60.16} & \textbf{60.16}  \\
$EN \rightarrow DE \rightarrow FR$ & 72.88 & 73.19 & \textbf{73.22} \\
$EN \rightarrow HI \rightarrow BN$ & 54.59 & 54.88 & \textbf{55.13}\\ 
\bottomrule
\end{tabular}
\caption{Performance evaluation of the proposed method using LLAMA2}
\label{table:decoder_results}
\end{table}

\begin{table*}[t!]
\centering
\small
\begin{tabular}{lp{0.7\linewidth}}
\toprule
\textbf{Component} & \textbf{Content} \\
\midrule
\multicolumn{2}{l}{\textbf{A. Prompt Template}} \\
\midrule
Template & \texttt{[INST] <<SYS>>\textbackslash n\{system\_message\}\textbackslash n<</SYS>>\textbackslash n\textbackslash n\{text\}\textbackslash n[/INST] [IN:\{label\}]} \\
\midrule
\multicolumn{2}{l}{\textbf{B. Instantiated Example}} \\
\midrule
\texttt{\{system\_message\}} & You are a task-oriented classification model. Output the top-level intent tag (e.g., [IN:TRAVEL\_BOOKING]). \\
\texttt{\{text\}} & book a flight from hyderabad to delhi for tomorrow \\
\texttt{\{label\}} & \texttt{TRAVEL\_BOOKING} \\
\midrule
\multirow{4}{*}{\textbf{Final Prompt (Training)}} & \texttt{[INST] <<SYS>>} \\
& \texttt{You are a task-oriented classification model. Output the top-level intent tag (e.g., [IN:TRAVEL\_BOOKING]).} \\
& \texttt{<</SYS>>\textbackslash n\textbackslash nbook a flight from hyderabad to delhi for tomorrow\textbackslash n[/INST] [IN:TRAVEL\_BOOKING]} \\
\midrule
\textbf{Final Prompt (Inference)} & ...`book a flight from hyderabad to delhi for tomorrow\textbackslash n[/INST] [IN:` \\
\bottomrule
\end{tabular}
\caption{The prompt structure for intent classification. Section A shows the general template with placeholders. Section B provides a concrete example, followed by the final formatted prompts used for model training and inference.}
\label{tab:prompt_structure}
\end{table*} 

\textbf{Limitations} Linguistic justification for why a specific POS category outperforms other POS categories. Although our approach is parameter-efficient, the exploration of other PEFT methods, including Prefix Tuning and QLoRA, may yield further improvements. To support evaluation of low resource languages, we translated samples using the NLLB model. While this improves multilingual coverage, translation quality may vary across languages.

\section{Conclusion}
This paper proposes a novel continual learning framework for adapting large language models (LLMs) to low-resource languages using adapter-based modular architectures and POS-based code switch exploiting the syntactic cues of the languages. To mitigate catastrophic forgetting in multilingual settings, we introduce a shared replay adapter that retains cross-lingual knowledge during sequential training. The proposed method augments replay data using a POS-based code switch, which replaces tokens in target language inputs with syntactically aligned counterparts from prior languages. Experiments across multilingual tasks including intent classification, question answering, and vision-language understanding demonstrate that POS-guided replay significantly improves retention and generalization, outperforming random and no-replay baselines. The approach scales effectively for continual multilingual model adaptation.


%
\bibliography{name}
\appendix
\section{Additional Experimental Setup Details}
\subsection{Software, Computational Resources and Training Time}
We use HuggingFace Transformers \footnote{\url{github.com/huggingface/transformers} version 3.4.0 pre-trained on 104 languages, including all languages covered in our evaluation.}, PyTorch\footnote{\url{https://pytorch.org}}, spaCy\footnote{\url{https://spacy.io}}, and AdapterHub\footnote{\url{https://adapterhub.ml}} for model implementation and preprocessing.
All experiments were conducted on an NVIDIA RTX A6000 GPU with 48 GB of VRAM. For continual training with a language sequence length of 3, MTOP completed in approximately 3 hours, PAXQA required 3.5 hours, while XQGA incurred a significantly higher runtime of 47 hours, attributed to the computational demands of the XMDETR architecture. We observe that the runtime difference between random and POS-based code-switching configurations remains within 10 minutes per experiment.
  
\subsection{Hyperparameter Sensitivity}
We perform hyperparameter sensitivity analysis by varying batch size, code switch ratio, and training epochs on the MTOP test dataset using our proposed method. Table \ref{table:hyperparameter} reports the average accuracy across configurations for the language sequence $EN \rightarrow HI \rightarrow BN$ using our proposed method.

\subsection{Dataset Properties}
The properties of the MTOP, PAXQ, and XGQA datasets are summarized in Tables \ref{table:mtopdataset}, \ref{table:paxqadataset}, and \ref{table:xgqaqadataset}, respectively. While MTOP and PAXQ are primarily used for training and validation, XGQA serves as a multilingual evaluation benchmark. To enable training on XGQA using the XMDETR architecture, we leverage the GQA dataset which is in English and apply code-switching techniques to incorporate non-English languages, following the multilingual adaptation strategy outlined in XMDETR \citep{Wang_2023_CVPR}. 

\begin{table}[t]
\small
\begin{tabular}{ccc}
\toprule
Language&	Train  (Approx.)	&Test  (Approx.) \\
\toprule
English  &	11,830	&4,386 \\
German &	9,575	&3,549 \\
Spanish &	8,085	&2,998 \\
French &	8,620	&3,193 \\
Hindi &	7,525	&2,789 \\
Thai &	7,460	&2,765 \\
\toprule

\end{tabular}

\caption {MTOP dataset languages and data splits}
\label{table:mtopdataset}
\end{table}

\begin{table}[t]
\small
\begin{tabular}{ccc}
\toprule
Language&	Train  (Approx.)	&Test  (Approx.) \\
\toprule
English  &	660,000	&1,788 \\
Arabic & 141,000 & 1,788 \\
Chinese & 334, 000 & 1,788 \\
Russian & 185,000 & 1,788 \\
\toprule

\end{tabular}

\caption {PAXQA dataset languages and data splits}
\label{table:paxqadataset}
\end{table}

\begin{table}[t]
\small
\begin{tabular}{ccc}
\toprule
Language & Validation/Test (Approx.) \\
\toprule
English 	&11,000 \\
German 	& 11,000 \\
Portuguese  &	11,000 \\
Russian &	11,000 \\
Indonesian  &	11,000 \\
Bengali & 		11,000 \\
Korean  &		11,000 \\
Chinese &	 	11,000 \\
\toprule
\end{tabular}

\caption {XGQA dataset languages and data splits}
\label{table:xgqaqadataset}
\end{table}

\begin{table}[t]
\small
\begin{tabular}{ccc}
\toprule
Hyperparameter & Value & Average Accuracy \\
\toprule
\multirow{3}{*}{Batch Size} &	16	&91.80 \\
 &	32	&89.08 \\
 & 	64&	86.94 \\
 \hline
\multirow{4}{*}{Epochs}	&3&	85.61 \\
	&5&	88.05 \\
	&7&	89.92 \\
	&10&	91.80 \\
\hline
\multirow{4}{*}{CS Ratio}	&0.1	&91.62\\
	&0.3	&91.72\\
	&0.5	&91.80\\
	&0.7	&91.57\\

\toprule

\end{tabular}

\caption {Hyperparameters and Average Accuracy}
\label{table:hyperparameter}
\end{table}

\section{Additional Experiments}

\subsection{Code Switch Base Language Analysis}
In this section, we investigate the choice of the base language in code-switching for experience replay. Departing from the conventional use of English as the default base, we explore alternative high-resource languages such as German and French to assess their impact on cross-lingual knowledge retention in continual learning on MTOP dataset. Table \ref{table:base_lang} shows that code switch based on the German language outperforms code switch based on the French language. However, code switch based on English (Table \ref{table:main_results}) performs better than German and French.

\begin{table*}[t] 
\small
\centering
\setlength{\tabcolsep}{2.3pt}

\begin{tabular}{ccccccccc}
\hline

\hline

 Language Sequence &  Base Language   & Random CS  &   {NOUN } &  {VERB} &  {ADJ} &{ADV} &  {PROPN} &  {INTJ}    \\ 
 \toprule
\multirow{2}{*}{$EN \rightarrow FR \rightarrow ES$} &  DE & 88.79 & 89.23 & 89.12 & 88.92 & 89.46 & 89.45 & 89.1  \\
&FR &66.74 & 65.74 & 64.14 & 66.22& 66.38 & 66.37 & 65.9 \\ 
\multirow{2}{*}{$EN \rightarrow ES \rightarrow FR$} & DE & 89.05 & 89.38 & 89.37 & 89.52 & 89.18 & 89.85 & 88.99 \\
& FR & 65.91 & 67.95 & 67.45 & 67.45 & 68.25 & 68.61 & 68.59 \\
\multirow{2}{*}{$EN \rightarrow TH \rightarrow KO$} & DE & 89.80 & 90.24 & 89.77 & 90.06 & 90.19 & 89.89 & 90.14 \\
& FR & 88.30 & 88.02 & 87.53 & 88.11 & 88.46 & 88.52 & 89.26 \\
\multirow{2}{*}{$EN \rightarrow DE \rightarrow FR$} & DE & 90.82 & 90.02 & 90.4 & 90.58 & 90.65 & 90.37 & 90.70  \\
& FR & 54.62 & 58.29 & 58.93 & 55.40 & 56.35 & 55.94 & 55.68 \\
\multirow{2}{*}{$EN \rightarrow HI \rightarrow BN$} & DE & 91.7 & 91.59 & 91.5 & 91.75 & 91.17 & 91.4 & 91.85   \\
& FR & 90.88 & 91.08 & 90.04 & 90.34 & 90.81 & 90.94 & 90.93 \\
\toprule

\end{tabular}
\\[5pt]
\caption {Analysis of base languages for code switch}
\label{table:base_lang}
\end{table*}

\subsection{Replacing Adapters with low-rank adaptation (LoRA) Modules in the Proposed Architecture
}
In this section, we evaluate our proposed method by substituting adapters with LoRA modules. We keep all hyperparameters the same as in the adapter setup, including setting the LoRA rank to 16. As shown in Table~\ref{table:lora}, the proposed method that uses adapter-based configuration consistently outperforms LoRA across all language sequences.

\subsection{Evaluating the Proposed Method under Language Sequence and Length Variation
 }
The Table \ref{table:main_results} presents results for language sequences of fixed length 3. To further assess the robustness of our method, we evaluate continual learning performance across varying sequence lengths on MTOP datset. The Table \ref{table:seq_len} demonstrates that POS-based curriculum strategies consistently outperform random curriculum baselines. Notably, as the language sequence length increases, we observe a gradual decline in average accuracy, indicating the growing challenge of knowledge retention and transfer in longer task sequences.

\begin{table*}[t] 
\small
\centering
\setlength{\tabcolsep}{2.3pt}

\begin{tabular}{ccccccccc}
\hline

\hline

 Language Sequence   & Random CS  &   {NOUN } &  {VERB} &  {ADJ} &{ADV} &  {PROPN} &  {INTJ} & Proposed Method   \\ 
 \toprule
          
 $EN \rightarrow FR \rightarrow ES$ & 57.00 & 57.07 & 57.03& 56.3 & 57.82 & 56.97 & 57.06 &\textbf{89.69}    \\

$EN \rightarrow ES \rightarrow FR$ & 58.87 & 59.98 & 60.3 & 60.85 & 59.88& 60.32 & 60.92 &  \textbf{90.07}        \\

$EN \rightarrow TH \rightarrow KO$   &59.61 & 59.63 & 59.56 & 59.68 & 59.45 & 59.33 & 59.51 &  \textbf{90.08}    \\

$EN \rightarrow DE \rightarrow FR $   &  66.36 & 65.17 & 64.09 & 65.46 & 64.82 & 65.52 & 65.4 & \textbf{90.63} \\

$EN \rightarrow HI \rightarrow BN$     &  83.6 & 83.75 & 83.51 & 83.82 & 83.78 & 83.4 & 83.74&  \textbf{91.8}  \\
 
\toprule

\end{tabular}
\\[5pt]
\caption {Comparison of LoRA and adapter setup}
\label{table:lora}
\end{table*}

\begin{table*}[t] 
\small
\centering
\setlength{\tabcolsep}{2.3pt}

\begin{tabular}{cccccccc}
\hline

\hline

 Language Sequence   & Random CS  &   {NOUN } &  {VERB} &  {ADJ} &{ADV} &  {PROPN} &  {INTJ}    \\ 
 \toprule
          
 $EN \rightarrow FR \rightarrow ES \rightarrow DE $ & 89.01 & 89.02 & \textbf{89.03} &  88.98 & 88.95 & 89.12 & \textbf{89.03} \\
 
 $EN \rightarrow FR \rightarrow ES \rightarrow DE \rightarrow HI$ & 84.65 & 85.21 & \textbf{85.59} & 85.13 & 84.49 & 85.00 & 85.2 \\

$EN \rightarrow FR \rightarrow ES \rightarrow DE \rightarrow HI \rightarrow BN$ &  79.12 & 79.01 & 79.35 & 79.41 & 79.32 & 79.15 & \textbf{79.93}  \\
 $EN \rightarrow TH \rightarrow KO \rightarrow BN $ & \textbf{88.49} & 88.34 & 88.26 & 88.21 & 88.29 & 88.26 & 88.33  \\
 $EN \rightarrow TH \rightarrow KO \rightarrow BN \rightarrow HI $ & 88.42 & \textbf{88.43} & 88.09 & 88.19 & 88.42 & 88.22 & 88.23 \\
$EN \rightarrow TH \rightarrow KO \rightarrow BN \rightarrow HI \rightarrow DE  $ & 88.44 & 88.38 & 87.89 & \textbf{88.55}& 88.28& 88.32 & 88.39 \\ 
 
\toprule

\end{tabular}
\\[5pt]
\caption {Evaluation of proposed method on different language sequence lengths}
\label{table:seq_len}
\end{table*}

\end{document}